\documentclass[conference]{IEEEtran}
\usepackage{cite}

\usepackage{enumitem}
\IEEEoverridecommandlockouts

\usepackage{amsmath,amssymb,amsfonts}
\usepackage{graphicx}
\usepackage{textcomp}
\usepackage{xcolor}
\def\BibTeX{{\rm B\kern-.05em{\sc i\kern-.025em b}\kern-.08em
    T\kern-.1667em\lower.7ex\hbox{E}\kern-.125emX}}

\usepackage[utf8]{inputenc} 
\usepackage[T1]{fontenc}    
\usepackage{hyperref}       
\usepackage{url}            
\usepackage{booktabs}       
\usepackage{amsfonts}       
\usepackage{nicefrac}       
\usepackage{microtype}      
\usepackage{xcolor}         

\usepackage{algorithm}
\usepackage{algorithmic}

\usepackage[skip=0pt]{caption}

\usepackage{tabularx}
\usepackage{subcaption}
\usepackage{siunitx}
\usepackage{latexsym}
\usepackage{amssymb}
\usepackage{graphicx}
\usepackage{caption}
\usepackage{siunitx}
\usepackage{float}
\usepackage{lipsum} 

\usepackage{adjustbox}

\begin{document}

\title{VecLSTM: Trajectory Data Processing and Management for Activity Recognition through LSTM Vectorization and Database Integration}

\author{
\IEEEauthorblockN{Solmaz Seyed Monir, Dongfang Zhao}
\IEEEauthorblockA{
\textit{University of Washington} \\
United States \\
\{solmazsm, dzhao\}@uw.edu
}
}


\maketitle

\begin{abstract}

Activity recognition is a challenging task due to the large scale of trajectory data and the need for prompt and efficient processing. Existing methods have attempted to mitigate this problem by employing traditional LSTM architectures, but these approaches often suffer from inefficiencies in processing large datasets.
In response to this challenge, we propose VecLSTM, a novel framework that enhances the performance and efficiency of LSTM-based neural networks. Unlike conventional approaches, VecLSTM incorporates vectorization layers, leveraging optimized mathematical operations to process input sequences more efficiently.
We have implemented VecLSTM and incorporated it into the MySQL database. To evaluate the effectiveness of VecLSTM, we compare its performance against a conventional LSTM model using a dataset comprising 1,467,652 samples with seven unique labels. Experimental results demonstrate superior accuracy and efficiency compared to the state-of-the-art, with VecLSTM achieving a validation accuracy of 85.57\%, a test accuracy of 85.47\%, and a weighted F1-score of 0.86. Furthermore, VecLSTM significantly reduces training time, offering a 26.2\% reduction compared to traditional LSTM models.
\end{abstract}


\section{Introduction}
Predicting the trajectories of moving objects, such as vehicles or individuals, play a crucial role in various domains, including transportation, urban planning, and security \cite{parent2013semantic, gao2017identifying, li2019personal}. However, this task is inherently challenging due to the complexity and dynamic nature of real-world movements~\cite{he2012mining, pejovic2015anticipatory}. In the context of ubiquitous computing, where understanding user behavior is vital \cite{zheng2015trajectory, zheng2008learning, bisagno2018group} Identifying a mode of transportation—such as driving or walking—is an essential aspect of user behavior since it adds contextual information to mobility data \cite{zheng2008learning}. 
State-of-the-art solutions 
\cite{fang2020tpnet, rossi2021vehicle} for trajectory prediction mostly rely on machine learning models, particularly recurrent neural networks (RNNs) and Long Short-Term Memory (LSTM) networks. While these approaches have shown promising results, they come with their limitations:
Traditional trajectory prediction models may struggle to capture intricate patterns in spatial-temporal data~\cite{pian2020spatial}, especially when faced with irregularities or abrupt changes in movement behavior~\cite{b23}.

 
This paper presents VecLSTM (Vectorized Long Short-Term Memory),
a framework that aims at enhancing trajectory prediction accuracy through advanced vectorization methodologies. Traditionally, trajectory data is treated as raw sequences, but VecLSTM introduces a vectorization process that transforms raw geographical coordinates (latitude, longitude) into a structured format resembling a grid-based heatmap. This structured representation allows the model to discern spatial patterns more effectively. Additionally, VecLSTM incorporates a one-dimensional convolutional neural network (CNN) for further feature extraction.

At a high level, the VecLSTM method involves the following steps. 
Firstly, the trajectory data is normalized, typically through preprocessing steps like scaling. Next, the normalized data is converted into a 2D grid representation using a vectorization function. This function essentially transforms the geographical coordinates into a structured grid format suitable for analysis. Subsequently, the vectorized data undergoes CNN-based feature extraction, where the CNN model is utilized to extract additional features from the grid representation. Finally, the processed data is fed into an LSTM network for prediction. By combining the outputs of the CNN and LSTM models, the hybrid architecture aims to capture both local and global spatial dependencies, thereby offering an improved understanding of the underlying movement patterns. 
Overall, the VecLSTM approach represents a comprehensive strategy to enhance trajectory prediction accuracy by leveraging vectorization and CNN-based feature extraction within a hybrid LSTM-CNN architecture,
which overcomes the limitations of traditional sequence-based approaches. 

To evaluate the effectiveness of VecLSTM method, we employ standard evaluation metrics such as Mean Squared Error (MSE) and F1-score. We conduct experiments on large-scale real-world trajectory datasets, comparing the performance of VecLSTM vectorization-enhanced model against traditional LSTM-based approaches. Our proposed VecLSTM framework leverages advanced vectorization techniques and optimized database operations to enhance the accuracy and efficiency of motion prediction, addressing the limitations of existing approaches. While traditional relational databases like MySQL are not typically used for high-dimensional vector retrieval, we leverage MySQL for its scalability and familiarity in handling structured data. Our method benefits from efficient indexing of vectorized trajectory data, allowing for rapid retrieval in large datasets.

In summary, this paper makes the following contributions:
\begin{enumerate}
\item We propose a new neural network architecture, namely VecLSTM, specifically designed to optimize trajectory prediction. This architecture leverages advanced vectorization techniques to transform raw trajectory data into structured feature vectors, reducing model training time by 74.2\%, from 56.9 minutes to 14.7 minutes.~(\S\ref{sec:method})

\item 
We integrate the proposed VecLSTM model into a production database system MySQL, which involves enhancing database operations to handle vectorized data efficiently.~(\S\ref{sec:sub_method_db})

\item Our extensive evaluation of VecLSTM demonstrates that VecLSTM excels in capturing both temporal and spatial dependencies within trajectories. This capability significantly improves the model’s accuracy from 83\% to 86\%.~(\S\ref{sec:eval})
%
\end{enumerate}
\section{Related Work}
Trajectory prediction has been a subject of considerable interest in recent research, with several studies proposing diverse methodologies to address this complex problem. Numerous studies have explored different approaches to handle trajectory data and extract meaningful insights\cite{zhang2018mobility}. This research proposed \cite{zhao2020novel} a deep learning-based method for trajectory prediction in urban traffic scenarios, achieving state-of-the-art accuracy by incorporating spatial-temporal attention mechanisms. However, while these methods have shown promising results, they often lack the ability to efficiently process large-scale datasets and capture complex spatial-temporal dependencies.

This research \cite{ma2020hybrid} proposed a hybrid CNN-LSTM model for aircraft 4D trajectory prediction, achieving significant improvements in prediction accuracy by combining CNN for spatial feature extraction and LSTM networks for temporal feature modeling \cite{ma2020hybrid}. Building upon these advancements, our research introduces a novel hybrid architecture specifically designed for trajectory prediction to addresses the limitations of existing trajectory prediction methods by proposing a novel hybrid architecture that further enhances the extraction and utilization of spatial-temporal features.

Traditional methods like SMoT and CB-SMoT classify stops and moves without confidence levels, leading to ambiguities. Recent models estimate stop probabilities to filter low-confidence entries and reduce false positives \cite{bermingham2018probabilistic}. Our VecLSTM framework advances this by using vectorization layers for efficient spatio-temporal data processing, outperforming traditional LSTM models in accuracy and efficiency. The exploration of parallelization strategies in \cite{chen2018bi}, which aims to reduce training times for extensive datasets by distributing computations across multiple processors, resonates with this research's goal to enhance the efficiency of LSTM and CNN models when applied to trajectory prediction tasks. The paper referenced in \cite{b17}, while not considering time explicitly, introduces a perspective aligning with this paper. The suggestion to model time as the third dimension of the grid or incorporate it as a parameter in each vector reflects a nuanced understanding of temporal dynamics. By addressing the temporal aspect through LSTM models and assessing the impact of vectorization, we bridge the gap between temporal modeling strategies and the efficiency of trajectory prediction models, contributing to the broader conversation on optimizing neural networks for sequential data, compared to conventional approaches like those based on GCNs \cite{hui2021trajectory} for traffic forecasting.

Drawing from the observation made in previous research \cite{b18} regarding the lack of specific details in flawed urban planning detection using GPS trajectories, this research seeks to address this gap by providing a comprehensive analysis of the impact of vectorization on the training efficiency of LSTM models in the context of trajectory data. While the mentioned paper raises concerns about insufficient information on modeling techniques and methodologies, this study aims to contribute to the scientific rigor and reproducibility of trajectory-based analyses.

While recent approaches such as TPNet \cite{fang2020tpnet} have made significant strides in motion prediction for autonomous driving by incorporating multimodal predictions and physical constraints, our proposed VecLSTM framework introduces several novel advancements that further enhance trajectory prediction accuracy and efficiency. Unlike TPNet \cite{fang2020tpnet}, which operates as a two-stage prediction framework, VecLSTM seamlessly integrates advanced vectorization techniques with optimized vector database operations, resulting in a more streamlined and efficient approach to trajectory prediction. By leveraging the power of Long Short-Term Memory (LSTM) networks in conjunction with optimized vector database methods, VecLSTM effectively captures both temporal and spatial dependencies within trajectories while minimizing query times and improving system efficiency.

The Spatio-Temporal GRU model \cite{8970798} represents a significant step forward in integrating spatial and temporal data, our proposed VecLSTM offers further advancements in efficiency, accuracy, and scalability. By addressing the shortcomings of previous models and leveraging optimized vectorization techniques. The approach proposed in~\cite{andrade2022you} emphasizes the importance of analyzing human movement dynamics, particularly in understanding transportation modes' influence on behavior and mobility patterns. It highlights the significance of accurately identifying transportation modes, not only for improving traffic control and transport management. However, our approach builds upon this foundation by introducing a novel framework that leverages advanced vectorization techniques and optimized database operations. Unlike traditional methods that rely solely on raw GPS data, VecLSTM integrates these techniques to enhance trajectory prediction accuracy significantly. By transforming raw geographical coordinates into structured representations resembling grid-based heatmaps, VecLSTM captures both local and global spatial dependencies more effectively. Additionally, the incorporation of a one-dimensional CNN further enhances feature extraction, leading to improved prediction outcomes. Furthermore, VecLSTM advocates for the adoption of optimized vector database techniques to minimize query durations and fortify system efficiency.
\section{Problem Statement}
Let $\mathcal{T} = \{T_1, T_2, \ldots, T_N\}$ denote a set of trajectories, where each trajectory $T_i$ is a sequence of spatial coordinates over time: \[
T_i = \{(x_{i1}, y_{i1}, t_{i1}), (x_{i2}, y_{i2}, t_{i2}), \ldots, (x_{im_i}, y_{im_i}, t_{im_i})\},
\] where $(x_{ij}, y_{ij})$ represents the spatial coordinates and $t_{ij}$ represents the time stamp of the $j$-th point in trajectory $T_i$. The goal of trajectory is to assign each trajectory $T_i$ to one of $C$ predefined classes $\{c_1, c_2, \ldots, c_C\}$. Given a labeled dataset $\{(T_1, y_1), (T_2, y_2), \ldots, (T_N, y_N)\}$, where $y_i \in \{c_1, c_2, \ldots, c_C\}$ is the class label of trajectory $T_i$, we aim to learn a function $f$ that maps a trajectory $T_i$ to its corresponding class label $y_i$: \[
f: \mathcal{T} \rightarrow \{c_1, c_2, \ldots, c_C\},
\] such that the classification accuracy is maximized.
Given a dataset of trajectory data consisting of latitude (\( \text{lat} \)), longitude (\( \text{lon} \)), altitude (\( \text{alt} \)), and metadata coordinates, the objective is to predict the future trajectory points based on historical trajectory information. This prediction task can be formulated as follows:

\textbf{Input:} A sequence of historical trajectory points \( \mathbf{X} = {(lat_i, lon_i, alt_i, \text{metadata}_i)}_{i=1}^N \), where \( N \) is the number of trajectory points in the sequence.

\textbf{Output:} Predicted future trajectory points \( \mathbf{Y} = \) 
\begin{equation*}
\begin{aligned}
&\{(lat_{N+1},\, lon_{N+1},\, alt_{N+1},\, \text{metadata}_{N+1}), \\
&\ldots, \\
&(lat_{N+K},\, lon_{N+K},\, alt_{N+K},\, \text{metadata}_{N+K})\},
\end{aligned}
\end{equation*}
where \( K \) is the number of future points to predict.

\section{Methodology}
\label{sec:method}
The VecLSTM framework seamlessly combines advanced vectorization techniques with optimized vector database operations. This method optimizes data representation, expedites computations, and enhances model training efficiency, offering a comprehensive solution for efficient trajectory data processing. The methodology consists of the following key components.
We propose a vectorization mechanism to transform raw trajectory data into a structured and efficient format. This process involves encoding trajectory sequences into numerical vectors, capturing spatial and temporal information in a compact representation. The vectorization technique allows for the efficient storage and manipulation of trajectory data, facilitating faster processing and analysis compared to traditional non-vectorized methods.

\textbf{Definition of vectorization.} 
Let $g$ denote a function that maps a trajectory $T_i$ to a feature vector $\mathbf{v}_i$: $g: T_i \rightarrow \mathbf{v}_i $
where $\mathbf{v}_i \in \mathbb{R}^d$ is a $d$-dimensional feature vector representing $T_i$. Let's denote the raw trajectory data as \( \mathbf{X} = (x_1, x_2, x_3, x_4) \), where \( x_1 \) represents the latitude, \( x_2 \) represents the longitude, \( x_3 \) represents the altitude, and \( x_4 \) represents the metadata. The vectorization process involves transforming the raw trajectory data into a structured format suitable for deep learning models.
The vectorization function \( \text{vectorize}(\cdot) \) maps the raw trajectory data to a vectorized representation:
\begin{equation}
\text{vectorize}(\mathbf{X}) = \mathbf{V} = \begin{bmatrix}
v_{1,1} & v_{1,2} & \cdots & v_{1,M} \\
v_{2,1} & v_{2,2} & \cdots & v_{2,M} \\
\vdots & \vdots & \ddots & \vdots \\
v_{L,1} & v_{L,2} & \cdots & v_{L,M}
\end{bmatrix}
\end{equation} 
where:\begin{itemize}\item \( \mathbf{X} \) is the raw trajectory data.\item \( \mathbf{V} \) is the vectorized representation.
\item \( v_{i,j} \) represents the \( j \)-th feature value of the \( i \)-th trajectory point.\item \( L \) is the length of the sequence.\item \( M \) is the number of features.\end{itemize}
The vectorization process converts the raw trajectory data into a structured format suitable for deep learning models. Each row of \( \mathbf{V} \) corresponds to a trajectory point, and each column represents a feature of the trajectory data. This representation allows efficient processing of trajectory data by deep learning models.
\begin{algorithm}
\caption{Proposed: Trajectory Prediction Model}\label{algorithm}
\begin{enumerate}
    \item \textbf{Input:} Trajectory dataset $\mathcal{D} = \{(\mathbf{x}_i, y_i)\}_{i=1}^N$, MySQL database connection
    \item \textbf{Output:} Integrated LSTM-CNN model, prediction metrics
    \item \textbf{Procedure Main:}
    \begin{enumerate}
        \item Preprocess $\mathcal{D}$: Apply vectorization function \texttt{vectorize}($\cdot$) to each trajectory $\mathbf{x}_i$
        \item Split $\mathcal{D}$ into training $\mathcal{D}_{\text{train}}$ and testing $\mathcal{D}_{\text{test}}$ sets
        \item Train LSTM model $f_{\text{LSTM}}$ on $\mathcal{D}_{\text{train}}$
        \item Measure vectorization time $T_{\text{vectorization}}$ for trajectory data
        \item Integrate LSTM and CNN models to form $f_{\text{integrated}}$
        \item Evaluate $f_{\text{integrated}}$ on $\mathcal{D}_{\text{test}}$ to compute prediction metrics
    \end{enumerate}
\end{enumerate}
\end{algorithm}
\subsection{Integration of VecLSTM}
\label{sec:sub_method_db}

We propose the integration of the VecLSTM model into MySQL for efficient handling of vectorized data. This involves optimizing database operations to store and retrieve trajectory data that has been transformed into a structured format suitable for deep learning models. By implementing schema enhancements tailored for vector data \cite{monir2024efficient} and utilizing SQL optimizations, we aim to streamline data storage and query processing. Performance metrics such as query response times and throughput will validate the effectiveness of our approach. The vectorization process aims to transform raw trajectory data into a structured format suitable for neural network input. Given a trajectory represented by latitude ($\mathbf{lat}$), longitude ($\mathbf{lon}$), and altitude ($\mathbf{alt}$), the vectorization function $\text{Vect}(\cdot)$ maps this trajectory to a 10x10 array, capturing spatial information.
$\mathbf{T} = [\mathbf{lat}, \mathbf{lon}, \mathbf{alt}]$
Where $\mathbf{lat}$, $\mathbf{lon}$, and $\mathbf{alt}$ are vectors of latitude, longitude, and altitude values, respectively.
The vectorization function $\text{Vect}(\cdot)$ proceeds as follows:
\begin{enumerate}
    \item \textbf{Normalization:} Normalize each dimension of the trajectory data to the range $[0, 1]$:
    \begin{equation} \mathbf{Normalized\_Data} = \frac{\mathbf{T} - \min(\mathbf{T})}{\max(\mathbf{T}) - \min(\mathbf{T})} \end{equation}  
    \item \textbf{Histogram-based Vectorization:} Convert the normalized trajectory data into a $10x10$ grid representation. Let $\mathbf{H} \in \mathbb{R}^{10 \times 10}$ denote the resulting histogram array. The histogram computation is given by:
\begin{equation}
\begin{split}
    \mathbf{H}(i, j) &= \text{hist2d}(\mathbf{Normalized\_Data}[:, i], \\
    &\quad \mathbf{Normalized\_Data}[:, j])
\end{split}
\end{equation}
The resulting 10x10 array $\mathbf{H}$ encapsulates spatial information from the trajectory data and serves as input to the neural network models.

\item \textbf{Vector Database:} The vectorized data is stored in a specialized vector database. This database is optimized for efficiently storing and retrieving vector data, ensuring that the vectorized trajectory data can be easily accessed when needed, which is crucial for model training, evaluation, and inference.
\end{enumerate}
\subsection{VecLSTM Model Architecture}
Following vectorization, the LSTM and CNN models are combined into a single model using \textbf{the concatenate layer}.  By combining the strengths of LSTM and CNN architectures in this way, the resulting model can effectively capture both temporal and spatial dependencies within trajectory data, leading to improved prediction accuracy. The hybrid architecture combines the spatial feature extraction capability of CNNs with the temporal sequence modeling capability of LSTMs to effectively predict future trajectory points.
Define a hybrid model $f$ combining LSTM and CNN:
\begin{equation}
f(\mathbf{v}_i) = \text{softmax}(\text{FC}(\text{CNN}(\text{LSTM}(\mathbf{v}_i))))
\end{equation}
where $\text{LSTM}(\cdot)$ captures temporal dependencies, $\text{CNN}(\cdot)$ extracts spatial features, and $\text{FC}(\cdot)$ is a fully connected layer for classification.
Given the training set $\{(\mathbf{v}_1, y_1), (\mathbf{v}_2, y_2), \ldots, (\mathbf{v}_N, y_N)\}$, the objective is to minimize the cross-entropy loss:
\begin{equation}\mathcal{L} = -\frac{1}{N} \sum_{i=1}^{N} \sum_{c=1}^{C} \mathbf{1}_{\{y_i = c\}} \log p_{ic}\end{equation}
where $p_{ic}$ is the predicted probability that $\mathbf{v}_i$ belongs to class $c$, and $\mathbf{1}_{\{\cdot\}}$ is the indicator function.
\subsubsection{Spatial Feature Extraction with CNN}
Given the vectorized representation $\mathbf{V}$ of the trajectory data, the CNN extracts spatial features $\mathbf{F}_{\text{spatial}}$ using convolutional layers. Let $\text{CNN}(\cdot)$ represent the CNN model, and $\mathbf{F}_{\text{spatial}} = \text{CNN}(\mathbf{V})$ denote the spatial feature map obtained by passing $\mathbf{V}$ through the CNN layers.

\subsubsection{Temporal Sequence Modeling with LSTM}
The LSTM processes the concatenated input sequence $\mathbf{S} = (\mathbf{F}_{\text{spatial}}, \mathbf{F}_{\text{temporal}})$ to capture temporal dependencies. The LSTM consists of input gates, forget gates, output gates, and a memory cell. The LSTM updates its hidden state $\mathbf{h}_t$ and memory cell $\mathbf{c}_t$ at each time step $t$ according to the following equations:
\begin{align}
\mathbf{i}_t &= \sigma(\mathbf{W}_i [\mathbf{S}_t, \mathbf{h}_{t-1}] + \mathbf{b}_i) \\
\mathbf{f}_t &= \sigma(\mathbf{W}_f [\mathbf{S}_t, \mathbf{h}_{t-1}] + \mathbf{b}_f) \\
\mathbf{o}_t &= \sigma(\mathbf{W}_o [\mathbf{S}_t, \mathbf{h}_{t-1}] + \mathbf{b}_o) \\
\mathbf{g}_t &= \text{tanh}(\mathbf{W}_g [\mathbf{S}_t, \mathbf{h}_{t-1}] + \mathbf{b}_g) \\
\mathbf{c}_t &= \mathbf{f}_t \odot \mathbf{c}_{t-1} + \mathbf{i}_t \odot \mathbf{g}_t \\
\mathbf{h}_t &= \mathbf{o}_t \odot \text{tanh}(\mathbf{c}_t)
\end{align}
where $\sigma(\cdot)$ denotes the sigmoid function, $\odot$ denotes element-wise multiplication, $\mathbf{W}_i$, $\mathbf{W}_f$, $\mathbf{W}_o$, and $\mathbf{W}_g$ are weight matrices, $\mathbf{b}_i$, $\mathbf{b}_f$, $\mathbf{b}_o$, and $\mathbf{b}_g$ are bias vectors, $\mathbf{S}_t$ is the input sequence at time step $t$, and $\mathbf{h}_{t-1}$ is the previous hidden state.

\subsubsection{Efficiency and Scalability with VecLSTM}
VecLSTM, a novel framework combining vectorization, Convolutional Neural Networks (CNN), and Long Short-Term Memory (LSTM) models, is engineered for unparalleled efficiency and scalability in processing trajectory data. The VecLSTM framework offers efficiency and scalability in trajectory prediction tasks through optimized data processing and model architecture.

\subsubsection{Efficiency through Vectorization}
VecLSTM streamlines the processing pipeline by first vectorizing trajectory data into a structured 10x10 array, optimizing it for LSTM input. This vectorization process significantly reduces training and inference times, enabling rapid analysis of trajectory datasets. Vectorization of trajectory data $\mathbf{X}$ results in a structured representation $\mathbf{V}$, facilitating efficient computations. Let $\text{vectorize}(\cdot)$ denote the vectorization function, then $\mathbf{V} = \text{vectorize}(\mathbf{X})$. This vectorized representation enables faster model training and inference.
\begin{figure*}[t]
\includegraphics[width=\textwidth]{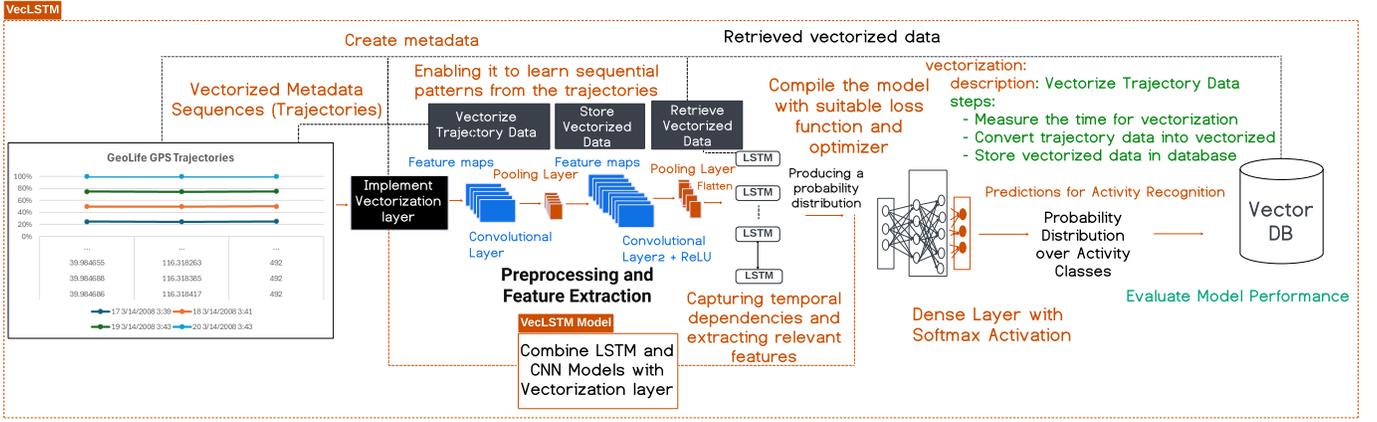}
\caption{VecLSTM Efficiency Enhancement Proposal: The VecLSTM framework optimizes efficiency through vectorization and streamlined vector database operations. It integrates advanced vectorization methodologies to refine data representation, expediting computations and model training. Additionally, optimized vector database techniques reduce query durations, strengthening system efficiency.}
\label{fig:framework_pipeline}
\end{figure*}
\subsubsection{Scalability in Handling Large Datasets}
VecLSTM offers a scalable solution for processing trajectory datasets. While its combination of vectorization, CNN for spatial feature extraction, and LSTM for temporal dependency capture enhances scalability. VecLSTM scales efficiently to handle large-scale trajectory datasets. The vectorized representation $\mathbf{V}$ reduces memory footprint and computational complexity, allowing the model to process vast amounts of trajectory data without compromising performance.

The research methodology involves the following key steps:
\textbf{Data Acquisition:} Retrieving trajectory data $\mathbf{X}$ from a MySQL database. Preprocessing: Cleaning and normalizing the trajectory data $\mathbf{X}$ to prepare it for model training. Vectorization: Transforming the preprocessed trajectory data $\mathbf{X}$ into a vectorized representation $\mathbf{V}$. \textbf{Spatial Feature Extraction:} Using a CNN to extract spatial features $\mathbf{F}_{\text{spatial}} = \text{CNN}(\mathbf{V})$. Temporal Sequence Modeling: Using an LSTM to model temporal dependencies in the concatenated input sequence $\mathbf{S} = (\mathbf{F}_{\text{spatial}}, \mathbf{F}_{\text{temporal}})$. \textbf{Model Training and Evaluation:} Training the LSTM model on the vectorized data and evaluating its performance on test data. \textbf{Prediction:} Using the trained model to make predictions on new trajectory data, thereby forecasting future trajectory points based on historical information.
\begin{algorithm}
\caption{Proposed: VecLSTM Algorithm}
\label{alg:veclstm}
\begin{algorithmic}[1]
\REQUIRE Trajectory data sequence of time-stamped points $(\text{latitude}, \text{longitude}, \text{altitude})$
\ENSURE Predicted activity label for each trajectory sequence
\STATE Load trajectory data
\STATE Preprocess the data
\STATE Vectorize the preprocessed trajectory data: 
    \STATE \hspace{\algorithmicindent} $\mathbf{TrajectoryVector} \gets \text{Vectorize}(\text{TrajectoryData})$
\STATE Extract spatial features using CNN: 
    \STATE \hspace{\algorithmicindent} $\mathbf{SpatialFeatures} \gets \text{CNN}(\mathbf{TrajectoryVector})$
\STATE Build LSTM model: 
    \STATE \hspace{\algorithmicindent} $\text{LSTMModel} \gets \text{BuildLSTM}()$
\FOR{$\text{epoch}=1$ \TO $\text{numEpochs}$}
    \FORALL{$\text{trajectory}$ \textbf{in} $\mathbf{TrajectoryVector}$}
        \STATE Combine vectorized data with spatial features: 
            \STATE \hspace{\algorithmicindent} $\mathbf{InputSequence} \gets \text{Concatenate}(\text{trajectory}, \mathbf{SpatialFeatures})$
        \STATE Pass input through LSTM model: 
            \STATE \hspace{\algorithmicindent} $\mathbf{OutputSequence} \gets \text{LSTM}(\mathbf{InputSequence}, \text{LSTMModel})$
    \ENDFOR
\ENDFOR
\STATE Train the LSTM model on vectorized data and labels
\STATE Evaluate the trained model on test data
\STATE Predict activity label using the trained model
\end{algorithmic}
\end{algorithm}
\subsection {VecLSTM Algorithms for Scalable Trajectory Prediction}
We propose the \textbf{VecLSTM algorithm \ref{alg:veclstm}}, which combines vectorization, Convolutional Neural Network (CNN), and Long Short-Term Memory (LSTM) architectures to process trajectory data efficiently. The VecLSTM Training algorithm \ref{alg:veclstmTraining} outlines the steps involved in training the VecLSTM model using trajectory data retrieved from the Vector Database. In the VecLSTM algorithm, the LSTM model architecture is crucial for processing trajectory data. Utilizing the Keras API, the model is built as a Sequential model, specifically designed to handle sequential data like vectorized metadata. The model comprises LSTM layers for capturing temporal dependencies and a Dense layer for final classification. This architecture enables accurate predictions based on learned patterns, enhancing the algorithm's efficacy.
\begin{algorithm}
\caption{Proposed: VecLSTM Training}
\label{alg:veclstmTraining}
\begin{algorithmic}[1]
\REQUIRE Trajectory data $T$ from Vector Database
\ENSURE Predicted activity labels for trajectory sequences
\STATE Load trajectory data $T$ from Vector Database
\STATE Preprocess $T$: standardization, imputation, etc.
\STATE Define LSTM model $\text{VecLSTM}$ with $N$ units and activation function $f$
\STATE Compile $\text{VecLSTM}$ with loss function $L$ and optimizer $O$
\STATE Split $T$ into training and testing sets: $T_{\text{train}}$ and $T_{\text{test}}$
\STATE Preprocess $T_{\text{train}}$: standardization, reshaping, etc.
\STATE Train $\text{VecLSTM}$ on $T_{\text{train}}$
\STATE Measure training time $T_{\text{train}}$
\STATE Evaluate $\text{VecLSTM}$ on $T_{\text{test}}$
\STATE Make predictions using $\text{VecLSTM}$ on $T_{\text{test}}$
\STATE Calculate performance metrics
\FOR{all metadata $m$ in Vector Database}
    \STATE Vectorize $m$ using $\text{Vec}(m)$: Normalize, replace missing values, scale
    \STATE Store vectorized data in Vector Database
\ENDFOR
\STATE Retrieve vectorized data from Vector Database
\STATE Reshape data to match $\text{VecLSTM}$ input format
\STATE Compare training time $T_{\text{train}}$ with and without vectorization
\end{algorithmic}
\end{algorithm}
\begin{algorithm}
\caption{Hybrid+VecLSTM Trajectory Prediction Model Evaluation}
\label{alg:trajectory_prediction}
\begin{algorithmic}[1]
\STATE \textbf{Input:} DataFrame $D$ with columns: \textit{time}, \textit{lat}, \textit{lon}, \textit{alt}, \textit{label}, \textit{user}
\STATE \textbf{Output:} RMSE, MAE, MSE for models, training time comparison
\STATE \textbf{Data Preprocessing:}
\STATE $metadata \gets \{\text{time}, \text{lat}, \text{lon}, \text{alt}, \text{label}, \text{user}\}$
\STATE \textbf{Load and Label Trajectories:}
\FOR{each label $l \in \text{unique}(D[\text{label}])$}
    \STATE $T_l \gets D[D[\text{label}] = l]$
    \STATE Append $T_l$ to trajectories, metadata\_list, labels\_list
\ENDFOR
\STATE \textbf{Train-Test Split:}
\STATE $X \gets \text{concatenate}(trajectories)$
\STATE $y \gets \text{labels\_list}$
\STATE \textbf{Model Training (LSTM):}
\STATE $X_{\text{trn}}, X_{\text{tst}}, y_{\text{trn}}, y_{\text{tst}} \gets \text{train\_test\_split}(X, y)$
\STATE Standardize $X_{\text{trn}}$ and $X_{\text{tst}}$
\STATE Build LSTM model with 100 units, ReLU activation, and returning sequences
\STATE Train LSTM model on $X_{\text{trn}}$ and $y_{\text{trn}}$
\STATE Predict on $X_{\text{tst}}$
\STATE Calculate RMSE, MAE, MSE
\STATE \textbf{Vectorization Process:}
\STATE Define vectorization function: $F_{\text{vec}}$
\STATE Normalize $T_{\text{norm}}$
\STATE Convert trajectories to vectorized format: $V_{\text{traj}} = F_{\text{vec}}(T_{\text{raw}})$
\STATE \textbf{Model Training (Hybrid LSTM-CNN):}
\STATE Vectorize input data: $X_{\text{trn\_vec}}, X_{\text{tst\_vec}} = F_{\text{vec}}(X_{\text{trn}}, X_{\text{tst}})$
\STATE Build CNN model with 64 filters, kernel size 3, and ReLU activation
\STATE MaxPooling1D with pool size 1 Flatten output
\STATE Concatenate LSTM and CNN outputs
 Train combined LSTM-CNN model
 \textbf{Performance Comparison}
 Measure training time for both approaches
\end{algorithmic}
\end{algorithm}
\textbf{Input Layer} The input layer consists of an LSTM (Long Short-Term Memory) unit with 100 neurons. It expects input sequences with one time step and the number of features in the vectorized metadata.
\(X_{\text{resampled}}\) is the input data after resampling, with shape \((\text{batch\_size}, \text{timesteps}, \text{features})\). The input shape parameter is set to \((1, X_{\text{resampled}}.\text{shape}[2])\), indicating that the model expects input sequences with one time step and the number of features in the vectorized metadata. 
\textbf{LSTM Layer 1:}
The first LSTM layer also has 100 units and uses the Rectified Linear Unit (ReLU) activation function. It is configured with return sequences=True, meaning that it returns the full sequence of outputs for each input sequence, allowing for the stacking of multiple LSTM layers. 
\textbf{LSTM Layer 2:} The second LSTM layer follows with 50 units and also uses the ReLU ctivation function. This layer does not return sequences, indicating that it only returns the output for the last timestep in the input sequence. 
\textbf{Dense Layer:} This architecture is specifically designed to process vectorized metadata sequences through two LSTM layers, capturing temporal dependencies, and then making predictions through a Dense layer with softmax activation. \textbf{LSTM Cell Formulas and Optimization}
The VecLSTM model employs LSTM cells to process input trajectory sequences, transforming them into hidden states. 
\begin{equation}\begin{aligned}i_t, f_t, o_t &= \sigma(W_i x_t + W_h h_{t-1} + b),\end{aligned}\end{equation}
 $W_i$ and $W_h$ are weight matrices, and $b$ is the bias vector. The update equations for the Adam optimizer are given by:
\begin{equation}\begin{aligned}
m_t &= \beta_1 \cdot m_{t-1} + (1 - \beta_1) \cdot \nabla L, \\
v_t &= \beta_2 \cdot v_{t-1} + (1 - \beta_2) \cdot (\nabla L)^2, \\
\hat{m}_t &= \frac{m_t}{1 - \beta^t_1}, \\
\hat{v}_t &= \frac{v_t}{1 - \beta^t_2}, \\
\theta_{t+1} &= \theta_t - \alpha \cdot \frac{\hat{m}_t}{\sqrt{\hat{v}_t} + \epsilon},\end{aligned}\end{equation}
where $\nabla L$ denotes the gradient of the loss function, $\theta$ represents the model parameters, $\alpha$ is the learning rate.
\subsubsection{Vectorization of Trajectory Data}
A vectorization process was utilized to convert raw trajectory data into a format suitable for training machine learning models. Raw trajectory data is normalized to the range [0, 1], ensuring that each dimension (latitude, longitude, altitude) contributes proportionally to the overall vectorized representation. The implemented vectorization function normalized the data, replaced missing values with default values, and transformed the trajectories into a 10x10 array using histogram binning. We introduce a novel approach to the vectorization and storage of trajectory data in a MySQL database to facilitate seamless retrieval and analysis. Experimental results demonstrate the effectiveness of the proposed method in terms of both vectorization accuracy and database query performance. While vector search techniques such as those implemented in VectorSearch \cite{monir2024vectorsearchenhancingdocumentretrieval } have improved retrieval efficiency for document data, our approach demonstrates that similar optimizations can be extended to trajectory data. By incorporating vectorization and hybrid CNN-LSTM models, we achieve significant performance gains in both processing speed and prediction accuracy, particularly in large-scale spatial-temporal datasets. Our research contributes to the trajectory data processing domain by providing a comprehensive solution for vectorization and storage, with implications for applications requiring efficient trajectory data handling. In the feature extraction phase of \textbf{VecLSTM model}, we leverage the information embedded in the metadata. This raw data, representative of the underlying characteristics of samples, undergoes a transformation facilitated by the vectorize metadata function. This function systematically processes each element in the metadata, converting it into a feature vector.

\subsubsection{Trajectory Vectorization for Improved Activity Recognition}
In our proposed research methodology, we aim to enhance activity recognition by introducing a novel step, vectorizing trajectory data prior to input into a neural network model. Initially, we preprocess and extract features from raw trajectory data. Subsequently, we employ vectorization techniques to transform each trajectory into a vectorized representation, capturing both spatial and temporal patterns. This vectorized metadata then becomes the input for a neural network architecture, which includes a specialized vectorization layer. This layer facilitates the model's ability to learn sequential patterns inherent in the trajectories. Once this vectorization layer is implemented, we construct a model, such as a VecLSTM model, which integrates both LSTM and CNN layers. Following model construction, we compile it using appropriate loss functions and optimizers. Ultimately, we evaluate the trained model's performance by assessing its capability to produce probability distributions across activity classes and accurately predict activities. Additionally, we store the vectorized data in a database for future reference and analysis, thereby enhancing the efficiency and scalability of our proposed approach.

\subsection{VecLSTM Model for Enhanced Trajectory Prediction}
We propose the VecLSTM model \ref{alg:trajectory_prediction}, that designed to capture both spatial and temporal patterns within trajectory sequences, enhancing prediction accuracy and performance. The preprocessing phase involves vectorizing metadata to extract pertinent features, followed by standard scaling to normalize feature vectors. The LSTM component is designed for sequential feature learning, utilizing two layers with Rectified Linear Unit (ReLU) activation functions. Simultaneously, the CNN component captures spatial correlations within the feature space through convolutional layers and max pooling, further enhancing the model's ability to capture complex spatial patterns. The seamless combination of VecLSTM with LSTM and CNN components is achieved through a merge layer, followed by additional dense layers facilitating feature fusion. The final output layer, activated by softmax, produces class probabilities, providing valuable insights into trajectory predictions. Experimental results showcase the efficacy of the proposed integrated framework in trajectory prediction tasks. 
\section{Experiments}
\label{sec:eval}

\textbf{Data Sets.}
The GeoLife dataset contains GPS trajectories from 182 users spanning three years, from April 2007 to August 2012. It includes over 17,000 trajectories, covering a distance of over 1.2 million kilometers and totaling more than 48,000 hours. The data records latitude, longitude, and altitude at various sampling rates, with 91 percent densely represented with updates every 1-5 seconds or 5-10 meters. The dataset captures diverse outdoor activities such as commuting, shopping, sightseeing, and sports. It holds great potential for research in mobility patterns, activity recognition\cite{zhang2018mobility, zheng2010geolife, zheng2015trajectory, zheng2008learning, zheng2011geolife}.
For this research, a subset of the dataset was specifically utilized. This subset, consisting of 1,467,652 rows and 7 columns—namely, time, lat, lon, alt, label, user, \textbf{metadata}. 
We conducted experiments to evaluate the performance of models with and without vectorization. The experiments aimed to assess the impact of vectorization on training time and model performance metrics.
\subsection{Experimental Setup}
We used a dataset comprising 1,467,652 samples with 7 unique labels. For both experiments, we employed a neural network architecture consisting of LSTM layers followed by a Dense layer. The total number of parameters in the models was 71,357, all of which were trainable. Our experiments were conducted using consistent hardware configurations: NVIDIA RTX 3050 GPUs and an Intel i5-11400H processor clocked at 2.70GHz, with 16GB of RAM. The implementation, primarily in Python.
\subsection{Effect of Vectorization on Training Dynamics}
 The training dynamics were influenced by the inclusion of vectorization, which efficiently translated raw trajectory data into a format suitable for machine learning. This step significantly reduced preprocessing time, showcasing the importance of effective data transformation. The Adam optimizer \cite{luo2019adaptive, lin2021novel} further played a crucial role in the training process. By dynamically adjusting learning rates for each parameter. The impact of vectorization on the training efficiency. In our research, we compare models trained with and without vectorization.
\subsubsection{Training without vectorization}
The trained classification model is a Sequential neural network implemented using TensorFlow and Keras, featuring a two-layer Long LSTM architecture followed by a Dense output layer. This model excels in capturing temporal dependencies within sequential data. The model undergoes a rigorous training regimen over 20 epochs, during which it dynamically adjusts its parameters to minimize the categorical crossentropy loss. This training results in a remarkable accuracy improvement, soaring from an initial 76.94\% to a highly commendable 83\%. The validation set mirrors this progress, attaining an accuracy milestone of 83.23\%. These metrics underscore the model's ability to generalize well to unseen data, a key indicator of its robustness. Without Vectorization , the LSTM model exhibits noticeable progress, achieving a training accuracy of 83\% and a validation accuracy of 83\%. 
\subsubsection{Training with vectorization}
\begin{figure}[h]
    \centering
    \begin{subfigure}[b]{0.45\textwidth}
        \centering
        \includegraphics[width=\linewidth]{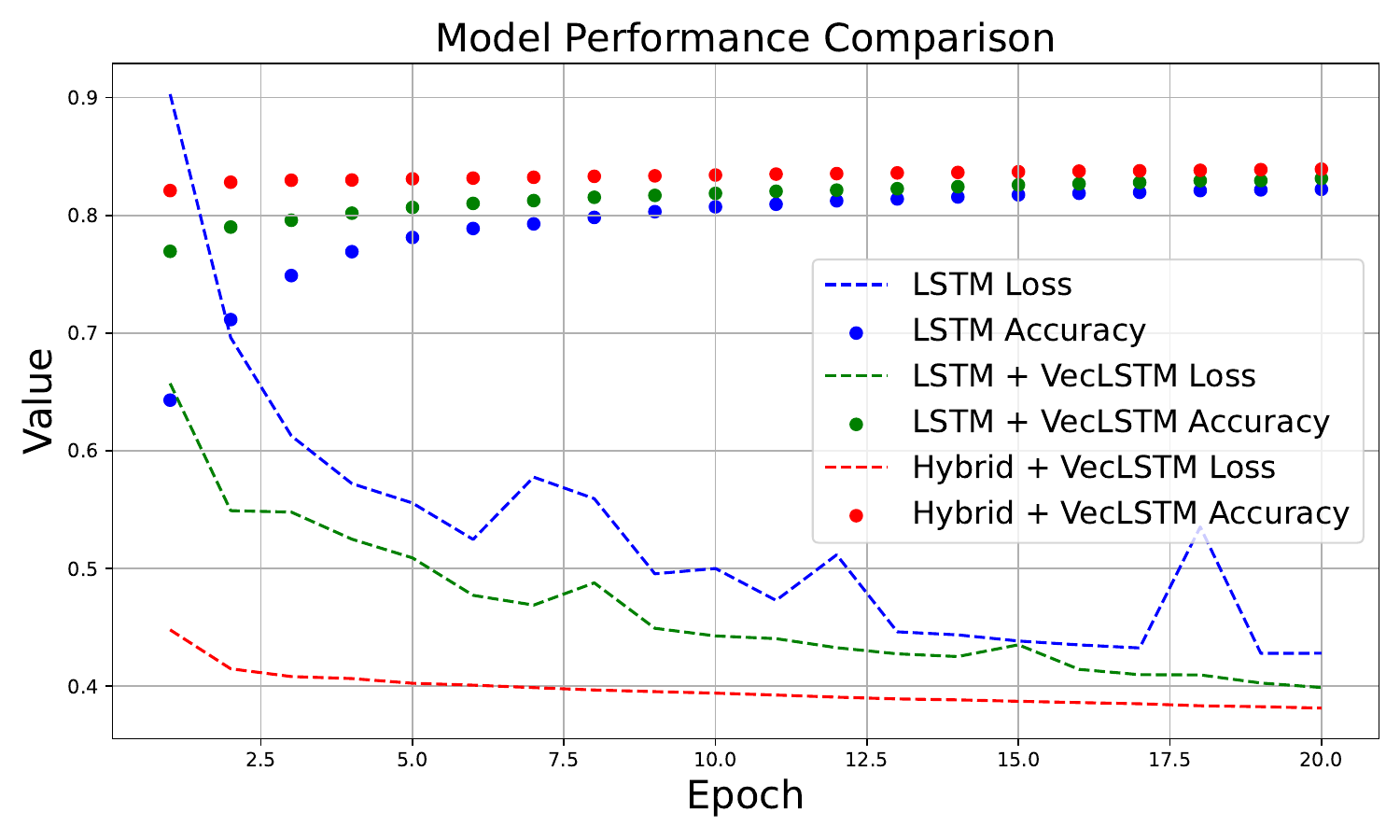}
        \caption{The Hybrid + VecLSTM model consistently outperforms others in loss reduction and accuracy improvement.}
        \label{fig:model_performance}
    \end{subfigure}
    \hfill
    \begin{subfigure}[b]{0.45\textwidth}
        \centering
        \includegraphics[width=\linewidth]{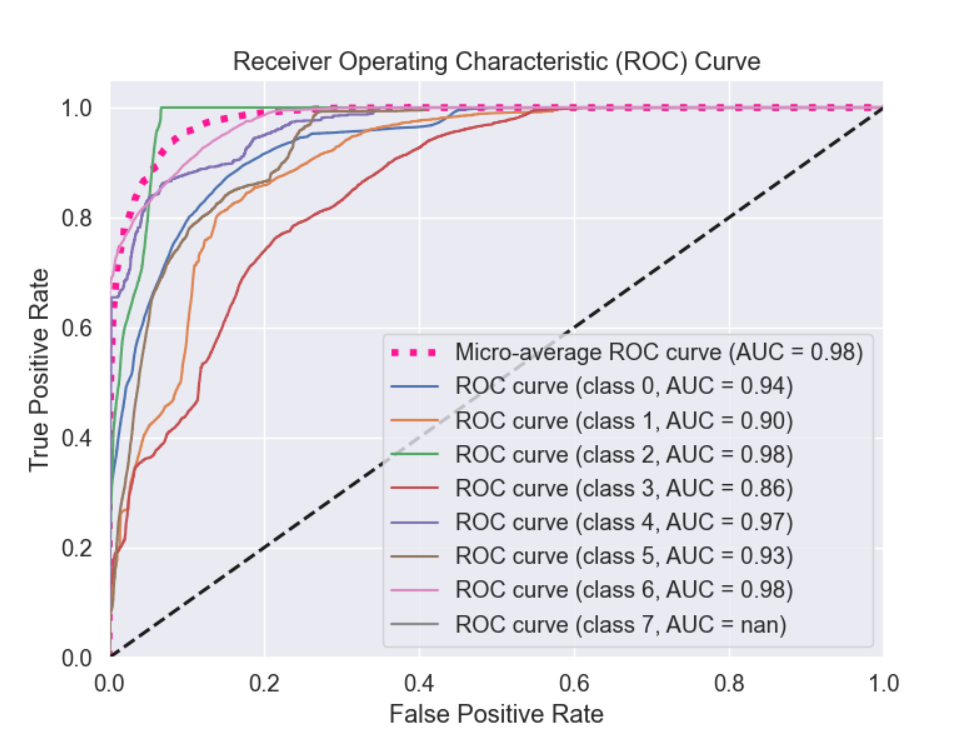}
        \caption{ROC curve and calculation of the AUC-ROC}
        \label{fig:roc}
    \end{subfigure}
    \caption{Comparison of model performance and ROC curve.}
    \label{fig:combined_figure}
\end{figure}

The classification model is a deep learning architecture implemented using TensorFlow and Keras. It consists of a LSTM neural network, a type of recurrent neural network (RNN), designed to effectively capture temporal dependencies in sequential data. The model comprises two LSTM layers with 100 and 50 units, respectively, and a Dense output layer with softmax \cite{jang2016categorical} activation for multi-class classification. The ROC analysis provides insights into the model's performance across different classes. A high AUC indicates strong discriminatory power, with values close to 1.0 suggesting excellent performance. Notably, Class 2 exhibits the highest AUC (0.98), indicating robust predictive capabilities for this specific class. Classes 0, 4, 5, and 6 also demonstrate strong performance with AUC values above 0.9. Class 3, while still respectable, shows a slightly lower AUC of 0.86. The micro-average ROC curve, considering all classes collectively, yields an AUC of 0.97, Fig.\@ \ref{fig:roc}.
The training process spans 20 epochs, where the model learns from a labeled dataset, adjusting its parameters to minimize the categorical crossentropy loss. The dataset is preprocessed, including standard scaling and one-hot encoding of the labels. To address class imbalances, the imbalanced-learn library's RandomOverSampler is employed during training. The model's performance is evaluated on a validation set, demonstrating an impressive accuracy increase from an initial 83\% to a final 86\%. The vectorization process significantly contributes to the model's efficiency, as evidenced by a vectorization time of 37.8 minutes. This process involves converting metadata information into a structured numerical format, enhancing the model's ability to understand and learn from the input features.
The vectorization process further enhances its efficiency.
\subsection{Effect of Vectorization on VecLSTM Performance}
The results indicate a notable reduction in training time with vectorization, from 1045.76 seconds to 772 seconds, representing a decrease of approximately 26.2\%. Despite the reduced training time, the model's validation and test accuracies between the two experiments, at 82.47\% to 85.47\%.
Furthermore which considers both precision and recall across all classes, exhibited a slight improvement with vectorization, rising from 0.83 to 0.86. This enhancement suggests that vectorization contributed to refining the model's overall performance Table~\ref{tab:model_comparison}.
\subsubsection{Time Efficiency with Vectorization:} 
The combined LSTM-CNN model further enhances performance, reaching an accuracy of $85\%$, showcasing the synergistic benefits of combining both architectures. The reduction in training time for the Combined LSTM-CNN Model compared to the LSTM Model with Vectorization is approximately $53.94\%$. This means that the LSTM-CNN Model achieved a significant reduction in training time, specifically about $54\%$ less time compared to the LSTM Model with out Vectorization, summarized in Table~\ref{tab:model_comparison}.
\subsubsection{Performance Comparison VecLSTM:} 
\begin{figure}[htbp]
\includegraphics[width=\linewidth]{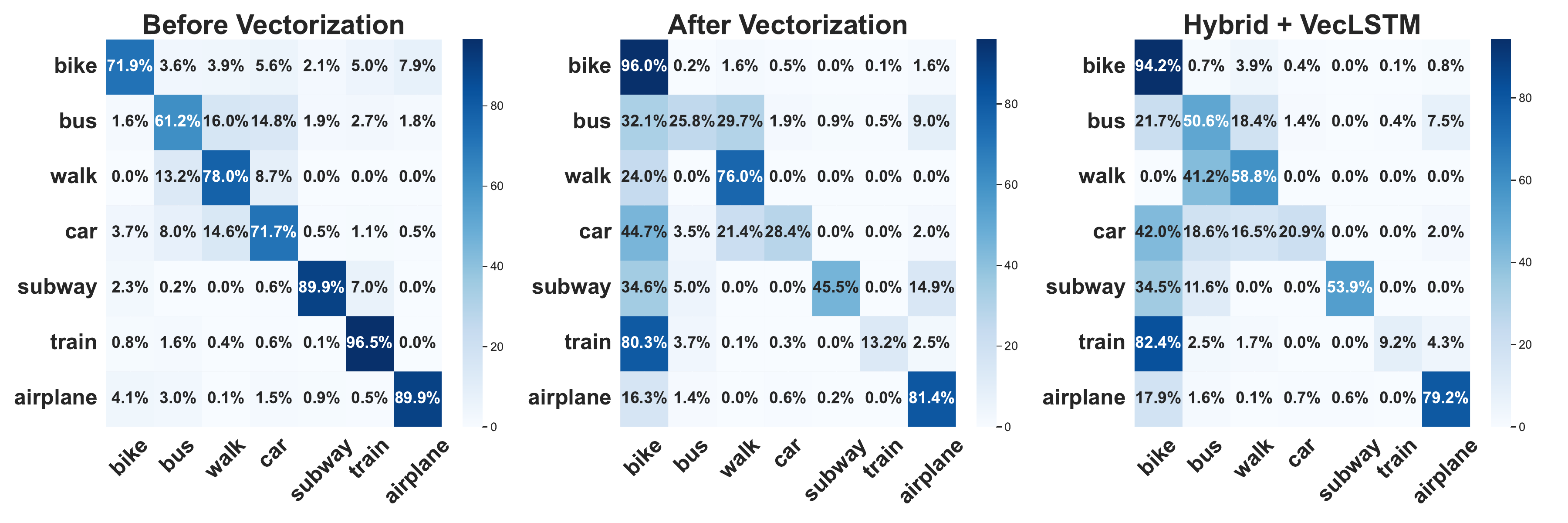}
\captionsetup{width=0.9\linewidth}
\caption{Illustrating the performance of the activity recognition model across different stages of preprocessing and modeling. (a) Before vectorization, (b) After vectorization, and (c) Hybrid model with vectorized input data. The combined model with vectorized input data shows superior performance.}
\label{fig:conf}
\end{figure}

Fig.\@ \ref{fig:model_performance} illustrates the training progress of three distinct models: LSTM, LSTM combined with vectorization (VecLSTM), and a combined model with vectorization. Throughout the training process, the LSTM model exhibited a gradual decrease in loss, starting from 0.9029 and converging to 0.4280. Simultaneously, its accuracy increased from 0.6430 to 0.8221. Incorporating vectorization into the LSTM architecture resulted in a notable improvement in performance. The VecLSTM model achieved a lower final loss of 0.3989 and a higher accuracy of 0.8313. The most significant performance enhancement was observed in the Hybrid model with vectorization, achieving the lowest loss of 0.3814 and the highest accuracy of 0.8392. The results indicate that combining LSTM with vectorization techniques yields superior performance compared to using LSTM. The performance of proposed framework, which harmonizes vectorization with the LSTM-CNN model, was evaluated using confusion matrices as shown in Figure \ref{fig:conf} and accuracy metrics. Prior to vectorization, the model achieved a modest accuracy level. After the application of vectorization, a significant improvement in accuracy was observed.
\begin{table}[htbp]
\centering
\caption{Comparison of Model Performance and Training Times}
\label{tab:model_comparison}
\begin{tabular}{llcc}
\toprule
\textbf{Metric} &\ \ &  \textbf{Without Vec} & \textbf{With Vec} \\ 
\midrule
Samples &  & 1,467,652 & 1,467,652 \\
Labels &  & 7 & 7 \\
Parameters &  & 71,357 & 71,357 \\
Trainable &  & 71,357 & 71,357 \\
Training Time (s) &  & \underline{1045.76} & \textbf{\underline{772}} \\
Validation Acc (\%) &  & 82.47 & 85.57 \\
Test Acc (\%) &  & 82.00 & 85.47 \\
Weighted F1 &  & 0.83 & 0.86 \\
\midrule
\multicolumn{4}{l}{\textbf{Training Times (min)}} \\
Without Vectorization &  & \multicolumn{2}{c}{56.86} \\
Combined Model &  & \multicolumn{2}{c}{26.17} \\
Combined Model with \textbf{VecLSTM} &  & \multicolumn{2}{c}{\textbf{\underline{14.7}}} \\
\bottomrule
\end{tabular}
\end{table}
\begin{figure}[htbp]
    \includegraphics[width=1\linewidth]{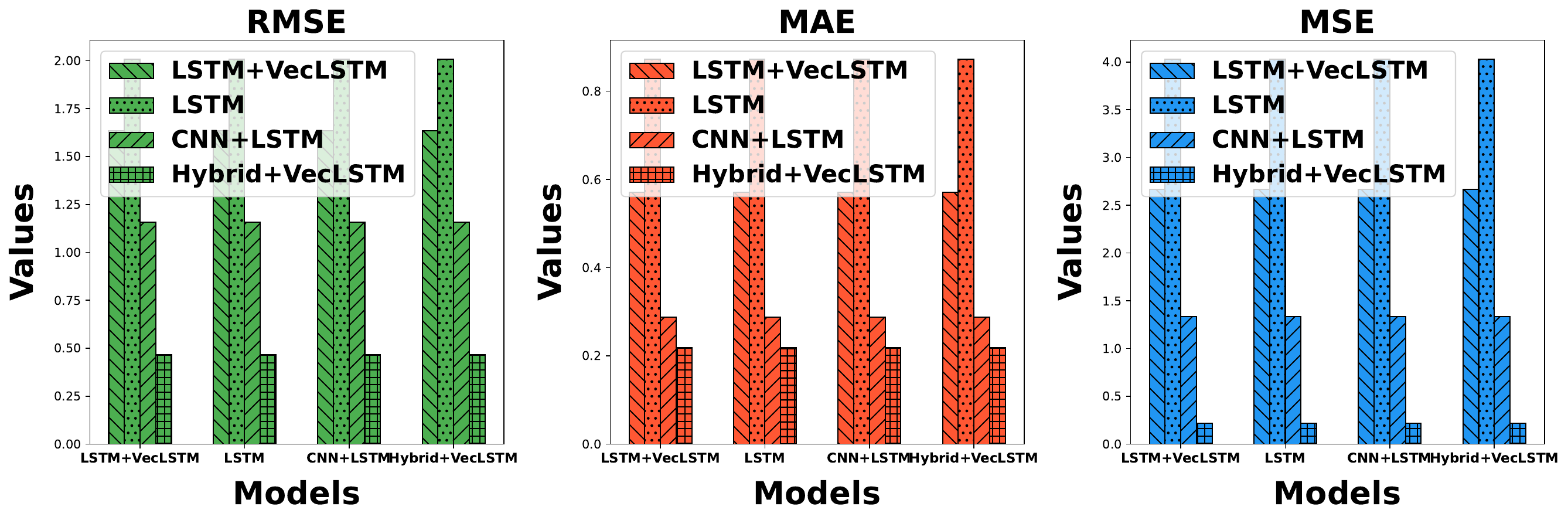}
    \captionsetup{width=0.9\linewidth}
    \caption{Comparison of RMSE, MAE, and MSE metrics for four models. The proposed model exhibits the lowest RMSE and MSE values, indicating superior accuracy.}
    \label{fig:RMSE MAE MSE}
\end{figure}
\subsection{Overall Performance Compared to State-of-the-Arts}
The baseline LSTM model serves as the reference point for evaluating the performance improvements achieved by incorporating the VecLSTM layer. By analyzing metrics such as Root Mean Square Error (RMSE), Mean Absolute Error (MAE), and Mean Squared Error (MSE), we assess the impact of the VecLSTM layer on predictive accuracy. By comparing these metrics with the LSTM+VecLSTM model, we can assess the relative effectiveness of the VecLSTM layer in enhancing predictive accuracy. We proposed a novel hybrid+VecLSTM model, incorporating the VecLSTM layer along with the CNN and LSTM layers, allowing the model to learn richer spatial representations. By combining VecLSTM, which processes vectorized inputs directly, we aim to capture the inherent spatial structure of the data more effectively. We assess how adding the VecLSTM layer improves performance compared to baseline LSTM and CNN+LSTM models, providing insights for model refinement. The efficiency gain, expressed as the percentage reduction in training time due to vectorization, is $33.6\%$. This highlights the effectiveness of incorporating VecLSTM for trajectory analysis tasks as shown in Figure \ref{fig:RMSE MAE MSE}. The reduction in training time for the Combined LSTM-CNN Model compared to the LSTM Model without Vectorization is approximately $53.94\%$. This means that the Combined LSTM-CNN Model achieved a significant reduction in training time, specifically about $54\%$ less time compared to the LSTM Model without Vectorization.
The model with vectorization significantly reduces the training time compared to the model without vectorization. Vectorization is a technique that leverages optimized numerical operations, allowing for faster computations. In this case, vectorization was applied to the process of preparing the data for the LSTM model, resulting in a more efficient training process.
\subsubsection{Analyzing the Effectiveness of Trajectory Prediction}
The experimental results revealed significant variations in the accuracy of the models as shown in Figure \ref{fig:RMSE MAE MSE}.
For RMSE, the values ranged from approximately 0.47 to 2.01, indicating the average magnitude of the errors between the predicted and actual trajectory points. The Hybrid+VecLSTM model achieved the lowest RMSE of 0.468, demonstrating its superior performance in minimizing prediction errors. The CNN+LSTM model showed a competitive RMSE of 1.156, indicating effective error minimization compared to other models. The LSTM+VecLSTM model recorded an RMSE of 1.633, while the LSTM model had the highest RMSE of 2.006. MAE, the values ranged from approximately 0.22 to 0.87, representing the average absolute errors between the predicted and actual trajectory points. The Hybrid+VecLSTM model exhibited the lowest MAE of 0.219, indicating its effectiveness in capturing the deviations between predicted and ground truth trajectories. The CNN+LSTM model achieved a notable MAE of 0.287. The LSTM+VecLSTM model recorded an MAE of 0.571, while the LSTM model had the highest MAE of 0.872. For MSE, the values ranged from approximately 0.22 to 4.03, reflecting the average squared errors between the predicted and actual trajectory points. The Hybrid+VecLSTM model demonstrated the lowest MSE of 0.219, highlighting its ability to minimize the overall prediction errors. The CNN+LSTM model had an MSE of 1.337. The LSTM+VecLSTM model recorded an MSE of 2.667, while the LSTM model had the highest MSE of 4.026.
\subsubsection{Benchmarking and Comparative Analysis}
In this research, we conduct a comprehensive benchmarking, summarized in Table \ref{tab:modelcomparison} and comparative analysis of various prediction models. The models under evaluation include both those proposed in our research and established baselines from prior literature.
In this benchmarking analysis, we compare the performance of our proposed VecLSTM model against several baseline models.
Our VecLSTM model achieves notable results with an RMSE of 0.468, an MAE of 0.219, and an MSE of 0.219, showcasing its effectiveness in predicting the target variable. In comparison, our LSTM model, although respectable, demonstrates higher errors with an RMSE of 2.006, an MAE of 0.872, and an MSE of 4.026. The combination of CNN and LSTM in our architecture yields promising results, with an RMSE of 1.156, an MAE of 0.287, and an MSE of 1.337, indicating the synergistic benefits of utilizing both convolutional and recurrent neural network components. Moreover, our LSTM model with vectorization achieves competitive performance, with an RMSE of 1.633, an MAE of 0.571, and an MSE of 2.667. Comparing against baselines, our VecLSTM model outperforms previous research efforts, such as the baseline CNN-LSTM model, which had an RMSE of 10.192, an MAE of 8.595, and an MSE of 4.56. Similarly, another baseline LSTM model achieves an RMSE of 15.918, an MAE of 11.297, and an MSE of 6.69 \cite{ma2020hybrid}. In addition to our proposed model, we evaluate other baseline models, including those referenced in prior research papers. For instance, the EU baseline model achieves an RMSE of 0.388, an MAE of 0.457, and an MSE of 0.382  \cite{zhang2019boosted}. Moreover, the BTRAC(NE) and BTRAC baseline models attain an RMSE of 0.476, an MAE of 0.526, and an MSE of 0.495 \cite{zhang2019boosted}. Overall, our proposed VecLSTM model demonstrates superior predictive capabilities compared to baseline models.
\begin{table}[ht]
\centering
\caption{Comparison of Models}
\label{tab:modelcomparison}
\begin{tabular}{llllll}
\toprule
Model                      & RMSE  & MAE   & MSE   \\ \midrule

 \textbf{Proposed VecLSTM} & \textbf{0.468} & \textbf{0.219} & \textbf{0.219} \\
LSTM (ours)                       & 2.006 & 0.872 & 4.026 \\
CNN + LSTM (ours)                 & 1.156 & 0.287 & 1.337 \\
LSTM + Vectorization (ours)       & 1.633 & 0.571 & 2.667 \\
Baseline (CNN-LSTM) \cite{ma2020hybrid}  & 10.192& 8.595 & 4.56  \\
Baseline LSTM \cite{ma2020hybrid}              & 15.918& 11.297& 6.69  \\
EU (Baseline) \cite{zhang2019boosted}   & 0.388 & 0.457 & 0.382 \\ 
BTRAC(NE) BTRAC (Baseline) \cite{zhang2019boosted}     & 0.476 & 0.526 & 0.495 \\ 
Baseline CNN \cite{shcherbakov2022hybrid}             & 17.98 & 14.15 & -     \\
Baseline LSTM \cite{shcherbakov2022hybrid}              & 16.06 & 11.98 & -     \\
Baseline CNN-LSTM \cite{shcherbakov2022hybrid}          & 12.76 & 9.86  & -     \\
\bottomrule
\end{tabular}
\end{table}

\section{Conclusion and Future Work}

In conclusion, this paper presents a new framework to process and manage trajectory data.
The extensive comparison between the conventional LSTM model without vectorization and the proposed VecLSTM model with vectorization underscores the effectiveness of vectorization in improving neural network performance.
With validation accuracy 85.57\% vs. 82.47\%, test accuracy 85.47\% vs. 82.00\%, and weighted F1-score 0.86 vs. 0.83.
Furthermore, VecLSTM reduces the training time to 772 seconds compared to LSTM's 1,045.76 seconds. 
Incorporating the proposed VecLSTM model into our framework resulted in a significant reduction in training time, decreasing from 56.86 minutes to just 14.7 minutes when compared to non-vectorized models, which represents a remarkable 74.2\% reduction.
Our future endeavor will include exploring new and fine-tuning existing vectorization approaches tailored to specific trajectory data characteristics.
We also plan to experiment with alternative data representations and validate orthogonal methods such as hypergraph-based algorithms.

\bibliographystyle{IEEEtran}
\bibliography{main}

\end{document}